\documentclass[10pt,twocolumn,letterpaper]{article}

\usepackage{wacv_zsl}
\usepackage{times}
\usepackage{epsfig}
\usepackage{graphicx}
\usepackage{amsmath}
\usepackage{amssymb}
\usepackage{times}
\usepackage{epsfig}
\usepackage{graphicx}
\usepackage{amsmath}
\usepackage{amssymb}
\usepackage{algorithm}
\usepackage{algorithmicx}
\usepackage[noend]{algpseudocode}
\usepackage{caption}
\usepackage{graphicx}
\usepackage{amssymb}
\usepackage{amsmath}
\usepackage{stmaryrd,mathtools}

\usepackage{varwidth}
\usepackage{makecell}
\usepackage{graphicx}
\usepackage{color}
\usepackage{subfig}
\usepackage{lipsum}
\usepackage{comment}
\usepackage{mathtools}

\DeclareMathOperator*{\argmax}{argmax}

\newcommand{\lbleq}[1]{\label{eq:#1}}

\newcommand{\ignorethis}[1]{}

\def\httilde{\mbox{\tt\raisebox{-.5ex}{\symbol{126}}}}

\newcommand{\norm}[1]{\lVert#1\rVert}

\wacvfinalcopy 

\def\wacvPaperID{622} 
\def\httilde{\mbox{\tt\raisebox{-.5ex}{\symbol{126}}}}

\ifwacvfinal\fi
\begin{document}

\title{A Generative Framework for Zero-Shot Learning  \\ with Adversarial Domain Adaptation}

\author{
Varun Khare$^1$\thanks{VK and DM contributed equally}, 
Divyat Mahajan$^2$\footnotemark[1] \thanks{DM and HB contributed while being part of IIT Kanpur}, 
Homanga Bharadhwaj$^3$\footnotemark[2], 
Vinay Kumar Verma$^1$,
Piyush Rai$^1$ 
\\  
$^1$IIT Kanpur 
$^2$Microsoft Research, India 
$^3$University of Toronto  \\
varunkhare1234@gmail.com, t-dimaha@microsoft.com, homanga@cs.toronto.edu \\ 
vkverma@iitk.ac.in , piyush@cse.iitk.ac.in}
\maketitle

\ifwacvfinal\thispagestyle{empty}\fi

\def\httilde{\mbox{\tt\raisebox{-.5ex}{\symbol{126}}}}


\begin{abstract}
  We present a domain adaptation based generative framework for zero-shot
 learning. Our framework addresses the problem of domain shift between the seen and unseen class distributions in zero-shot learning and minimizes the shift by developing a generative model trained via adversarial domain adaptation. Our approach is based on end-to-end learning of the class distributions of seen classes and unseen classes. To enable the model to learn the class distributions of unseen classes, we parameterize these class distributions in terms of the class attribute information (which is available for both seen and unseen classes). This provides a very simple way to learn the class distribution of any unseen class, given only its class attribute information, and no labeled training data. Training this model with adversarial domain adaptation further provides robustness against the distribution mismatch between the data from seen and unseen classes. Our approach also provides a novel way for training neural net based classifiers to overcome the hubness problem in zero-shot learning. Through a comprehensive set of experiments, we show that our model yields superior accuracies as compared to various state-of-the-art zero shot learning models, on a variety of benchmark datasets. Code for the experiments is available at github.com/vkkhare/ZSL-ADA
\end{abstract}

\section{Introduction}
In the conventional image classification tasks, examples from all classes are available during the training of the model. This assumption rarely holds in real-world problems, where we do not have the corresponding ubiquity of representative images from each class. Also, it is common knowledge that humans do not require prior visual evidence of a category to recognize an example from that category. Given that a child sees a picture of a horse and reads a description about zebra's appearance, he/she would more likely than not be able to easily recognize a zebra when an image is shown. The zero-shot learning (ZSL) problem~\cite{cmt,xian2018zero} in machine learning is motivated by similar considerations and seeks to exploit the existence of a labeled training set of `seen' classes and the knowledge about how each `unseen' class relates semantically to the seen classes.  

The success of ZSL lies in learning an effective semantic representation (e.g. attributes / textual features) for the successful transfer of knowledge from the seen to the unseen classes. In Sec.~\ref{sec:relwork}, we provide a detailed overview of the prior work on ZSL, but in particular generative ZSL methods~\cite{xian2018feature,wang2017zero,verma2017simple,vermageneralized} have become quite popular recently, by the virtue of their ability to \emph{generate} labeled examples for the unseen classes. However, a key requirement in such methods is the reliable estimation of the class distribution of seen and unseen classes. Even then, zero-shot learning suffers from hubness problem \cite{dem} mostly because of the use of nearest neighbor classifiers exploiting different distance metrics. It can be mitigated by using neural nets or any classifier which does not explicitly compare the inter-class distances in high dimensional data for label prediction. Hence, a generative model makes it plausible to train deep classifiers on synthesized data from the unseen classes. 

\begin{figure*}[!htbp]
    \centering
    \includegraphics[width=0.82\textwidth]{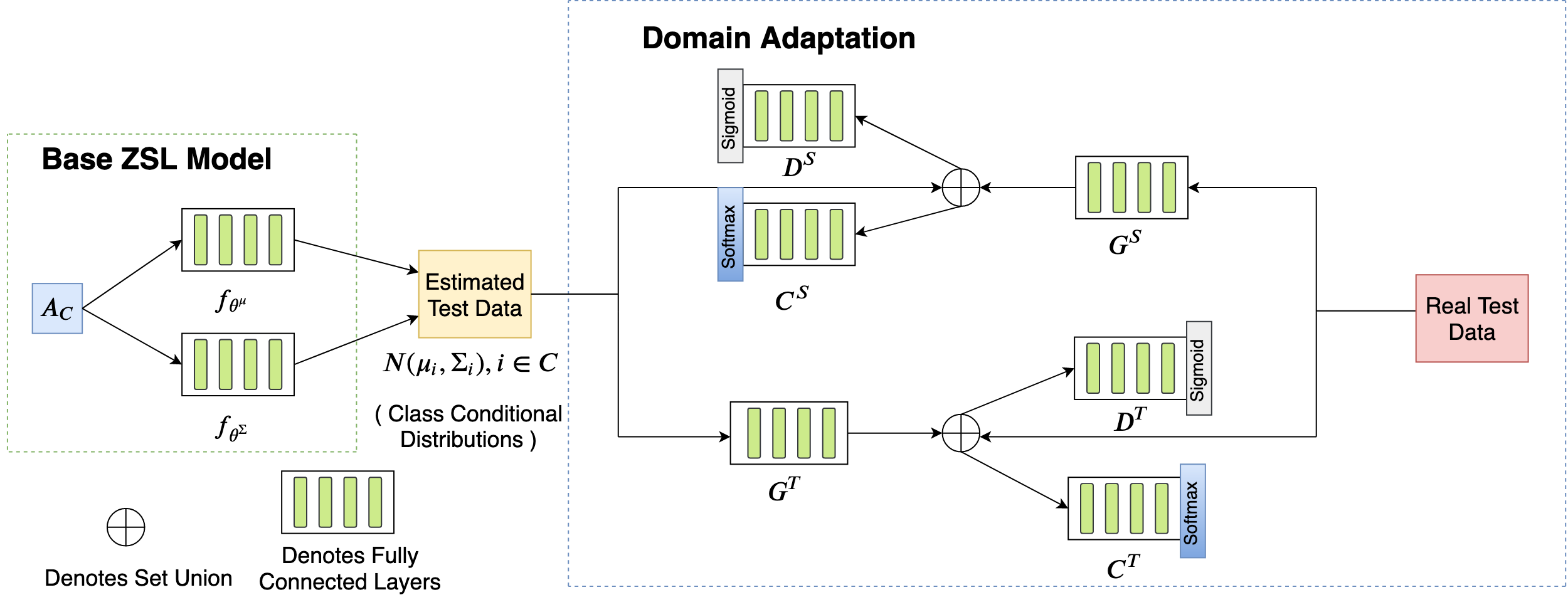}
    \caption{The overall architecture of the proposed approach. All the notations are consistent with that described in Section 2. $\mathbf{A_c}$ denotes the class attributes for all classes i.e. $\{\mathbf{a_c}\}_{c=1}^{S+U}$. }
    \label{fig:architecture}
\end{figure*}

 A simple, yet principled, way to construct generative models for ZSL is to learn the class distributions for the seen and unseen classes~\cite{verma2017simple,wang2017zero}. While this is straightforward for seen classes (for which we have access to labeled data), it can't be done for the unseen classes. In recent work, ~\cite{verma2017simple} used exponential family to model the distributions of the class conditionals in terms of learnable parameters. This is an effective model; however, their approach does not extend to non-exponential family distributions. Moreover, they used offline learning techniques to learn the parameters of seen classes, and rely on kernel-based regression to estimate the class parameters, given the class attributes. The model also requires careful tuning of hyperparameters. Our model, on the other hand, exploits the advantages of neural nets and end-to-end training to provide stability during the learning phase and remains less susceptible to hyperparameter variations.
 
However, such a model alone is not sufficient as there may be a domain shift between the original unseen class data and the synthesized unseen class data. The presence of acute domain shift between the seen and unseen classes hinders the performance of ZSL models~\cite{UDA}. Since the predictions for the unseen classes rely on the transfer of knowledge learnt from the seen classes, we might have poor performance on unseen classes due to the domain shift. We note that by enforcing domain adaptation to tackle problem settings where the train and test distributions are far apart, the model's performance can be greatly improved. An earlier approach~\cite{UDA} used the idea of \emph{joint} sparse coding for minimizing domain shift between the seen and unseen class data, however, since then there have been developments in adversarial domain adaptation that enable robust detection and resolution of domain shift~\cite{adda,cycada}. Adversarial learning and adaptation methods have found applicability in a wide range of fields from robotics navigation~\cite{icra} to recommender systems~\cite{recgan,irgan}. Several adversarial adaptation techniques like ADDA\cite{adda} require explicit source-target pairs of data points. Such a luxury is not present in Zero-Shot transductive setting where the test data is unlabelled. Similarly, unsupervised domain matching methods like CycleGAN\cite{cyclegan} use cyclic consistency to find the data point most similar to the source sample and then minimize the gap between these two. Though this is effective in maintaining the inherent clusters, it can match the unrelated class clusters together in the source and target domain if the classes are close enough.

Motivated by these desiderata, in this work, we develop an Adversarial Domain Adaptation framework for ZSL that leverages a generative ZSL model to improve upon the classification for unseen classes. Our model can transform the synthesized samples for unseen classes into the true test/unseen class domain while maintaining the data clusters associations. We first learn a generative model for the class conditional distribution of the seen and unseen classes by utilizing labeled data from the seen classes. Then, by domain adaptation, we explicitly bring closer the learnt distribution and the true distribution of the unseen class conditionals. We employ a scheme of cyclically consistent adversarial domain adaptation~\cite{cycada} to minimize domain shift without assuming any particular parametric form of the source and target distributions.

To the best of our knowledge, there is no adversarial framework for semi-supervised domain matching where explicit pairs of data points are not given but an external agent associates noisy labels to the samples. In addition, since we leverage neural nets for classification, we overcome the hubness issue, by virtue of not classifying based on explicit distances (unlike classical KNN type algorithms).

\section{The Proposed Approach}
\label{sec:model}

Our approach consists of a pre-training phase followed by adversarial domain adaptation (ADA). We first describe the generative model and then elaborate on the ADA setting. A detailed illustration of our method is shown in Figure 1.

\subsection{The Generative Model}
We model the data from class $c$ by a class conditional probability $p(\mathbf{x}|c,\mathbf{\Theta})$ where $\mathbf{\Theta}$ denotes the global parameters of the model. We do not have any restriction on the the type of distribution chosen. Let us denote the total number of classes whose examples are seen during training by $S$, and the classes, none of whose examples are seen during training by $U$. For the sake of defining the prediction rule formally, assume the unseen classes are known. Then, for an observation $\mathbf{x}$ from either a seen or unseen class $c$, where $c \in [1, S+U]$, we have $y_n = c$, and, assume the input to be generated as 
$
 x_n \sim p(\mathbf{x}|c,\mathbf{\Theta})
$ 

Under this framework, given test example $\mathbf{x_+}$, the predicted class $\hat{y}_+$ can be given by computing the most-probable class as follows
$
    \hat{y}_{+} = \argmax_c p(c|\mathbf{x_+},\mathbf{\Theta}) 
$ and using Baye's Rule we have,
\begin{equation}
    p(c|\mathbf{x_+},\mathbf{\Theta})  = \frac{p(\mathbf{x_+}|c,\mathbf{\Theta}) p(c|\mathbf{\Theta})}{\sum_{c \in [1,S+U]}{p(\mathbf{x_+}|c,\mathbf{\Theta}) p(c|\mathbf{\Theta}) }}
\end{equation}
Thus
\begin{equation}
    \hat{y}_+ = \underset{c}{\argmax}\  p(\mathbf{x_+}|c,\mathbf{\Theta})p(c|\mathbf{\Theta}) 
\end{equation}

For the sake of simplicity, we ignore the estimation of class prior probabilities and choose to treat them as equal for all the classes. However, their correct estimation can in principle provide better results. The prediction rule then becomes:
\begin{equation}
    \hat{y}_+ = \underset{\mathbf{c}}{\argmax}\  p(\mathbf{x_+}|c,\mathbf{\Theta})
\end{equation}
If labeled training data for all the classes are available, then standard inference techniques like Maximum Likelihood Estimation (MLE), Maximum-a-Posteriori (MAP) Estimation, or fully Bayesian inference can be used to determine the class conditional distributions. However, since the unseen classes do not have labeled training examples, we need a way to ``extrapolate'' the seen class distribution parameters to unseen class distribution parameters. This will be done via the class attribute vectors as described ahead (\ref{sec:attmapping}). 

First, assuming $\mathbf{X}$ and $\mathbf{C}$ ($c_k \in S\cup U$) denote the inputs and the associated output class labels respectively, a standard generative approach seeks to  maximize their joint distribution $\mathbb{P}[\mathbf{X^{S\cup U},\mathbf{C^{S\cup U}}}|\mathbf{\Theta}]$.

Assuming i.i.d. observations, we have
\begin{equation}
    \mathbb{P}[\mathbf{X^{S\cup U},\mathbf{C^{S\cup U}}}|\mathbf{\Theta}] = \mathbb{P}[\mathbf{X}^S,\mathbf{C^{S}}|\mathbf{\Theta}] \mathbb{P}[\mathbf{X^U},\mathbf{C^{U}}|\mathbf{\Theta}]
\end{equation}
Since, $\mathbf{X}^U,\mathbf{C^{U}}$ are unavailable for $\Theta$ estimation, 
usually $\mathbb{P}[\mathbf{X^{S},\mathbf{C^{S}}}|\mathbf{\Theta}]$ is maximized instead, expecting the learnt $\Theta$ to behave as a proxy to true value.
\begin{equation}
    \mathbb{P}[\mathbf{X^{S},\mathbf{C^{S}}}|\mathbf{\Theta}]=\prod_{\mathbf{x},c \sim S} p(\mathbf{x},c|\mathbf{\Theta}) 
\end{equation}
\begin{align}
   \Rightarrow log(\mathbb{P}[\mathbf{X^{S},\mathbf{C^{S}}}|\mathbf{\Theta}])=\underset{\mathbf{x},c \sim S}{\sum}log(p(\mathbf{x},c|\mathbf{\Theta})) \nonumber \\
   =\underset{\mathbf{x},c \sim S}{\sum}log(p(\mathbf{x}|c,\mathbf{\Theta}))+log(p(c|\mathbf{\Theta})) 
\end{align}
Since we are not modelling the class probability distribution $p(c|\mathbf{\Theta})$, the objective becomes
\begin{equation}
    \underset{\mathbf{\Theta}}{\argmax} \underset{\mathbf{x},c \sim S}{\mathbb{E}}[log(p(\mathbf{x}|c,\mathbf{\Theta}))]
\end{equation}

This sub-optimal $\Theta$ produces an inherent domain shift between the true unseen class distribution and the learnt distribution. We mitigate this by using adversarial domain adaptation to bring the unseen distribution and learnt distributions closer (refer \ref{sec:ADA}).

\subsubsection{Mapping Class Attributes to Class Parameters}
\label{sec:attmapping}
Since each class is described in terms of attribute vectors $\mathbf{a}_c$, we condition the class distribution on their respective attribute vector $\mathbf{a}_c$. Let these class-specific parameters be $\zeta_c$ which can be uniquely determined from the class attribute vector $\mathbf{a_c}$ and global parameters $\mathbf{\Theta}$ by a functional mapping $f$. This mapping for most purposes will be a complicated relationship and using a linear mapping (e.g., as done in~\cite{verma2017simple}) here would severely affect the generation quality of the network. We model this function $f:\{\mathbf{a}_c\} \rightarrow \{\mathbf{\zeta}_c\}$ using neural networks with trainable weights $\mathbf{\Theta}$  bringing extensive expressiveness and hierarchical relationships among attribute features. Thus, the class parameters can be written as  
\begin{equation}
    \mathbf{\zeta_c} = f_{\Theta}(\mathbf{a_c})
\end{equation}
\begin{figure}[H]
\centering
\begin{minipage}[c]{0.45\textwidth}
\centering
    \includegraphics[width=\textwidth,height=0.45\textwidth]{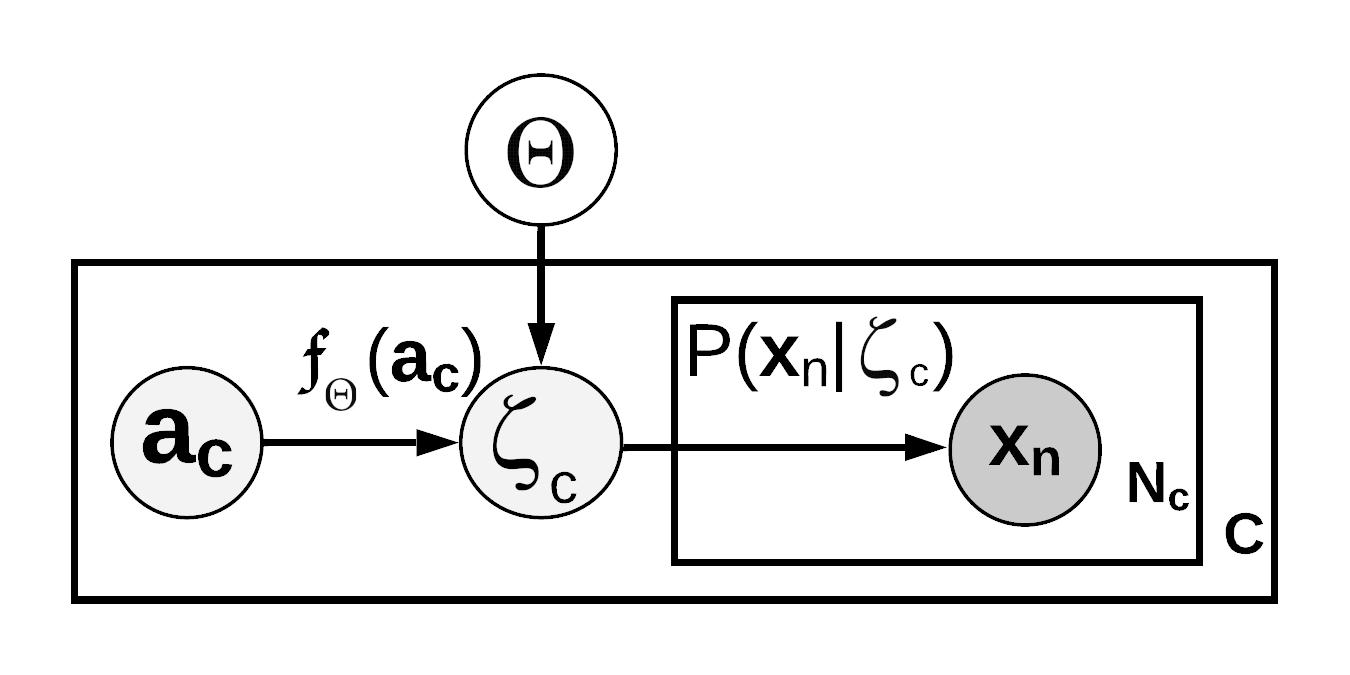}
    \caption{Samples of seen/unseen classes $\textbf{x}_n$ generated by the class conditional distribution defined by parameters $\zeta_c$  which in turn are the outputs of neural network $f_\Theta$ with $\textbf{a}_c$ as input}
    \label{fig:PGM}
\end{minipage}
\end{figure}

Such an approach leads to a stable training procedure w.r.t hyperparameters and enables us to perform the joint learning of $f_{\boldsymbol{\Theta}}$ and consequently, class parameters $\{\zeta_c\}$ in an end to end fashion. This is an important difference between our approach and the generative approach used by~\cite{verma2017simple}. Their approach first learns the class conditional parameters and then learns the attribute to class parameter mappings. We provide empirical justification for the stability of our approach in the Results section.

For simplicity, we take $p(\mathbf{x}|c,\mathbf{\Theta})$ to be a Gaussian distribution with parameters mean and co-variance, $\zeta_c =  \{\mathbf{\mu}_c,\Sigma_c \}$ where $c \in S$. 
We model $\mathbf{\mu_c}$ and $\mathbf{\Sigma}^{-1}_c$ as non-linear functions of the attribute vector $\mathbf{a_c}$ with neural networks of weights $\mathbf{\Theta}=\{ \mathbf{\theta^{\mu},\theta^{\Sigma}} \}$ in the following manner,
\begin{equation*}
    \mathbf{\mu_c} = f_{\theta^{\mathbf{\mu}}}(\mathbf{a_c}),\
        \mathbf{\Sigma_c}^{-1} = diag(f_{\theta^{\mathbf{\Sigma}}}(\mathbf{a_c})),\      \mathbf{x} \sim N \left(\mu_c, \mathbf{\Sigma_c}\right)
\end{equation*}
To ensure the condition of the covariance matrix ($\Sigma_c$) being a positive semi-definite matrix we model the inverse covariance to a diagonal matrix with positive diagonal entries. Thus $f_{\theta^{\Sigma}}$ outputs a vector in $\mathbb{R}^d_{>0}$ where d is the dimension of mean vector (equivalently the dimension of semantic space). The overall objective function becomes: 
\begin{align}
\small
\underset{\mathbf{\theta^\mu,\theta^\Sigma}}{\argmax} \underset{(\mathbf{x},c) \sim S}{\mathbb{E}}\left[log(\mathbf{\Sigma_c}^{-1} )
    -\left(\mathbf{x} - \mathbf{\mu_c}\right)^T  \mathbf{\Sigma_c}^{-1}\left(\mathbf{x} - \mathbf{\mu_c}\right)\right]
\end{align}

We again emphasize the fact that choosing Gaussian distribution is only for expositional purposes and one can also try other non-exponential family distributions as a part of inductive bias. The model does not restrict the choice of class conditional distribution to exponential family distributions
We take $\mathbf{x_n}$ as the features extracted from dataset images by resnet-101\cite{resnet} pre-trained on the Imagenet dataset \cite{imagenet2015}.  For the rest of the paper, $\{\mathbf{x}_n\}_c$ denotes the entire test data comprising of samples from all the classes. Similarly, $\{y_n\}_c$ denotes the samples generated from the generative model (as defined here) for all the classes. For the rest of the paper, we refer to the generative model defined above as the base ZSL model.


\subsection{Adversarial Domain Adaptation}
\label{sec:ADA}

The procedure described in the previous section only leverages the data from seen classes to estimate the class conditional distributions of all the classes. However, if there is a domain shift between the seen and unseen classes, then the estimated unseen class conditional distribution would also suffer from this domain shift due to reliance on the seen class data. Hence, to mitigate the issue of domain shift,  we propose to incorporate the unlabelled data from the unseen classes. In our overall architecture, we denote the process of learning the ZSL model parameters $\mathbf{\Theta}$ as `pre-training.' Based on the generative framework learned during pre-training, we can sample the class-conditional distribution for unseen classes to generate the unseen samples. We then minimize the domain gap between the generated distribution of the unseen classes and the actual distribution of the unseen classes. 

In this section, we denote the source domain as $S$ and the target domain as $T$. Through adversarial adaptation, we aim to bring the target distribution of $\{y_n \}_c$ (referred as $ y_{nc}$) closer to the source distribution  of $\{x_n \}_c$ (referred as $x_{nc}$); hence we learn a function $G^T(y_{nc})$ that maps class conditionals from the generated distribution $y_n$ to the real test distribution $x_n$ for all unseen classes $\{c\}_{c=1}^{U}$. Hence, $G^T: S\rightarrow T$ is a mapping from source $S$ to target $T$. Similarly, we define another function $G^S: T\rightarrow S$ that maps the class conditionals from the real test distribution to the same latent space as the class conditionals from the generated distribution. $D^T$ and $D^S$ are the corresponding discriminators.

Our design is inspired by CycleGAN \cite{cyclegan} and we make modifications to its base architecture for supporting zero shot learning. We consider a cyclic consistency loss instead of the vanilla adversarial loss (and variants) primarily because we want to learn as constrained a latent space for the Generator as possible. Cyclic consistency is an additional constraint on top of the adversarial loss that acts as an appropriate regularizer for transfer learning, as motivated in the original paper~\cite{cyclegan} . We also justify the cyclic consistency loss empirically in the Ablation Study section 4.4

\subsubsection{Label Augmentation}
Inspired by conditional GAN\cite{conGAN}, we augment the input to the generators $G^T$ and $G^S$ with the respective class labels, which facilitates the preservation of relationships between the synthesized data and their correct class labels. For $G^S$ the input data (test data) is unlabeled, hence we use the predictions from our pre-trained ZSL model as the guiding labels. Note that these labels are noisy labels and both the generators and discriminators should be capable of handling data corruption during the training phase. This is yet again a problem with the conventional GAN architectures. 
\subsubsection{Classifiers}
 We handled label recovery by adding two classifiers ($C^T,C^S$) in parallel with the discriminators. The parameters of the classifiers are trained jointly with the corresponding discriminators. Recently, parallel to our work, \cite{robustGAN} gave theoretical support to the use of external classifier with conditional GAN architecture to counter noisy data labels. The classifiers provide an additional benefit of enforcing linear separability for the generator but the impact is reduced if the classifier is multilayered. We justify clustering in the next section. 

\subsubsection{Optimization Function}
Let the loss defined in CycleGAN which consists of cyclic consistency loss ($\mathcal{L}_{cyc}$) and the adversarial loss ($\mathcal{L}_{adv}$) for domains T and S be $\mathcal{L}$
\begin{equation}
  \mathcal{L}= \mathcal{L}_{cyc}+\mathcal{L}_{adv}^T+\mathcal{L}_{adv}^S  
\end{equation}
For our case, $L_1$ norm worked the best for cyclic consistency loss\cite{cyclegan} $\mathcal{L}_{cyc}$, while Wasserstein loss \cite{arjovsky2017wasserstein} was found suitable for $\mathcal{L}_{adv}$. Additionally, identity regularizer with $l1$ norm (see eq 12) was added to the generator to ensure that the output domains for each generator remain unmodified if given as input. We add the classification loss ($\mathcal{L}_{clf}$) of the real data (not generated by G) to the discriminator loss during adversarial training. We ensured that the classification loss is not added at the beginning for data transformed by generators in accordance with mismatch loss addition for only real images~\cite{conGAN}. We evaluated cross entropy loss for both the correct and mismatched pairs of label-image. This enforces a stronger clustering than considering the loss term for only mismatched pairs, as in \cite{conGAN}. However once the GAN training has converged and the classification accuracy over the generated data becomes close or greater than the accuracy of pseudo-labels, we do a corruption recovery by training the classifiers over both the transformed samples $G^T(y_{nc})$ and true data samples $x_{nc}$ (\textit{refer eq (\ref{eq: classfifier})})

With $\chi, \xi,\beta$ as tune-able hyper parameters, the overall loss function then becomes
\begin{equation}
  \mathcal{L}= \mathcal{L}_{adv}^T+\mathcal{L}_{adv}^S+\chi \mathcal{L}_{cyc}+\xi\mathcal{L}_{clf}^T + \xi\mathcal{L}_{clf}^S  
\end{equation}
where $\mathcal{L}_{adv}^{\{T,S\}} = \{L_G +L_D\}^{\{T,S\}}$ with
\begin{equation}
    L_G^{T} = \underset{c \sim {p_{c}}}{\mathbb{E}}[\beta \norm{G^T(x_{nc})-x_{nc}}_p - D^T_w \circ G^{T}(y_{nc})]
\end{equation}
\begin{equation}
L^{T}_D =  \underset{c \sim {p_{c}}}{\mathbb{E}}[D^{T}_w \circ G^{T}(y_{nc})] - \underset{c \sim {p_{c}}}{\mathbb{E}}[D^T_w(x_{nc})] 
\end{equation}
Here, $D_w$ is the Wasserstein loss~\cite{arjovsky2017wasserstein} and $c$ denotes the class label.
\begin{align}
& \mathcal{L}_{\text{cyc}}(G^T, G^S)  =  \mathbb{E}_{c\sim p_{\text{c}}}[\norm{G^{S}\circ G^T(y_{nc})- x_{nc}}_p] \nonumber \\  &
+  \mathbb{E}_{c\sim p_{\text{c}}}[\norm{G^{T}\circ G^S(x_{nc})- y_{nc}}_p].\lbleq{cycle}
\end{align}
Here, $||\cdot||_p$ denotes the $L_p$ norm.
\begin{align}
L^{T}_{clf} = \mathbb{E}_{c \sim {p_{c}}}[L(C^{T}_{clf}\circ G^{T}(y_{nc}),Y^{T})] \nonumber \\  
+ \mathbb{E}_{c \sim {p_{c}}}[L(C^{T}_{clf}(x_{nc}),\bar{Y}^U)]
\label{eq: classfifier}
\end{align}
Similarly, we can define $L^S_{clf} , L^S_D , L^S_G$. Please refer to supplementary  material for exact equations and training algorithm.
\section{Related Work}
\label{sec:relwork}
Due to its ability to overcome the drawbacks of conventional classification problems, ZSL has attained tremendous recent interest for a wide range of AI problems, including those in computer vision. Earlier works~\cite{DAP,IAP} on ZSL were based on directly or indirectly mapping the instances of specific examples to their class-attributes. The learned mapping was then used during inference; this mapping works by first projecting the unseen data to the class-attribute space and then using the nearest neighbor search to classify the unseen image.
Another approach for ZSL focuses on learning the map of bi-linear compatibility between the visual space and the semantic space. ALE \cite{akata2013label}, DEVISE \cite{frome2013devise}, SJE \cite{SJE}, ESZSL \cite{ESZSL2015}, and SAE \cite{SAE2017} are based on the approach of measuring the bi-linear compatibility.


Generative models~\cite{mishra2017generative,Chen_2018_CVPR,xian2018feature,verma2017simple,guo2017synthesizing,BucherZSL,wang2017zero} have shown promising results for both ZSL and GZSL setups. Another advantage of the generative approach is that by using synthesized samples, we can convert the ZSL problem to the conventional supervised learning problem that can handle the biases towards the seen classes. The \cite{verma2017simple} used a simple generative model based on the exponential family framework while \cite{guo2017synthesizing} synthesize the classifier. While recent generative approaches for the ZSL are deep generative models based on the VAE \cite{VAE} and GAN \cite{GAN}. The approach \cite{vermageneralized,BucherZSL,mishra2017generative} is based on the VAE architecture while \cite{xian2018feature,Chen_2018_CVPR,lu2017zero} used the adversarial sample generation based on the class conditioned attribute. 

In ZSL, the train and test classes are disjoint and hence there is a high probability of domain shift for the unseen classes. This is another challenge in the ZSL setup and needs to be handled. Previously, very few works have handled the domain shift problem and worked on both the transductive as well as inductive settings. \cite{verma2017simple} adapted to the new domain by simple Gaussian mixture model updates. \cite{song2018transductive} used the unbiased embedding in the transductive setting. \cite{kodirov2015unsupervisedDA,ye2017zero_DOMAIN} proposed unsupervised domain adaption for the ZSL. \cite{saligram2016learningJoint} used the structural SVM formulation for domain adaption.


In this paper, we propose the design of a deep generative model that has many differences as compared to the previously proposed VAE/GAN based deep generative models for ZSL. The VAE based architecture minimizes the ELBO \cite{VAE} by using a scheme of approximate optimization, making it less robust in handling domain shift. This also applies to the latent class distributions learned by VAE. While we explicitly estimate the class conditional distributions, VAE based methods learn these distributions as latent variables via approximate inference. Thus the complexity of our model in representing the class conditionals is on par with VAE based models but on the other hand, we reap the benefits of direct optimization. The GAN based generative approach is difficult to train, requiring a lot of seen class examples during training. Moreover, they need the attribute vectors of unseen classes at the beginning of the procedure while our model can handle on the fly addition of new classes.
To this end, we propose a simple CNN based architecture that can learn any parametric distribution with exact optimization, and unlike the GAN based approach, has stable training. This makes it especially suitable for domain shift minimization by adapting the distribution of the unseen classes. 

\section{Experiments and Results}
\label{sec:model}
\begin{figure*}[t]
    \centering
    \begin{tabular}{ccc}
        \includegraphics[width=0.32\textwidth,height=0.30\textwidth]{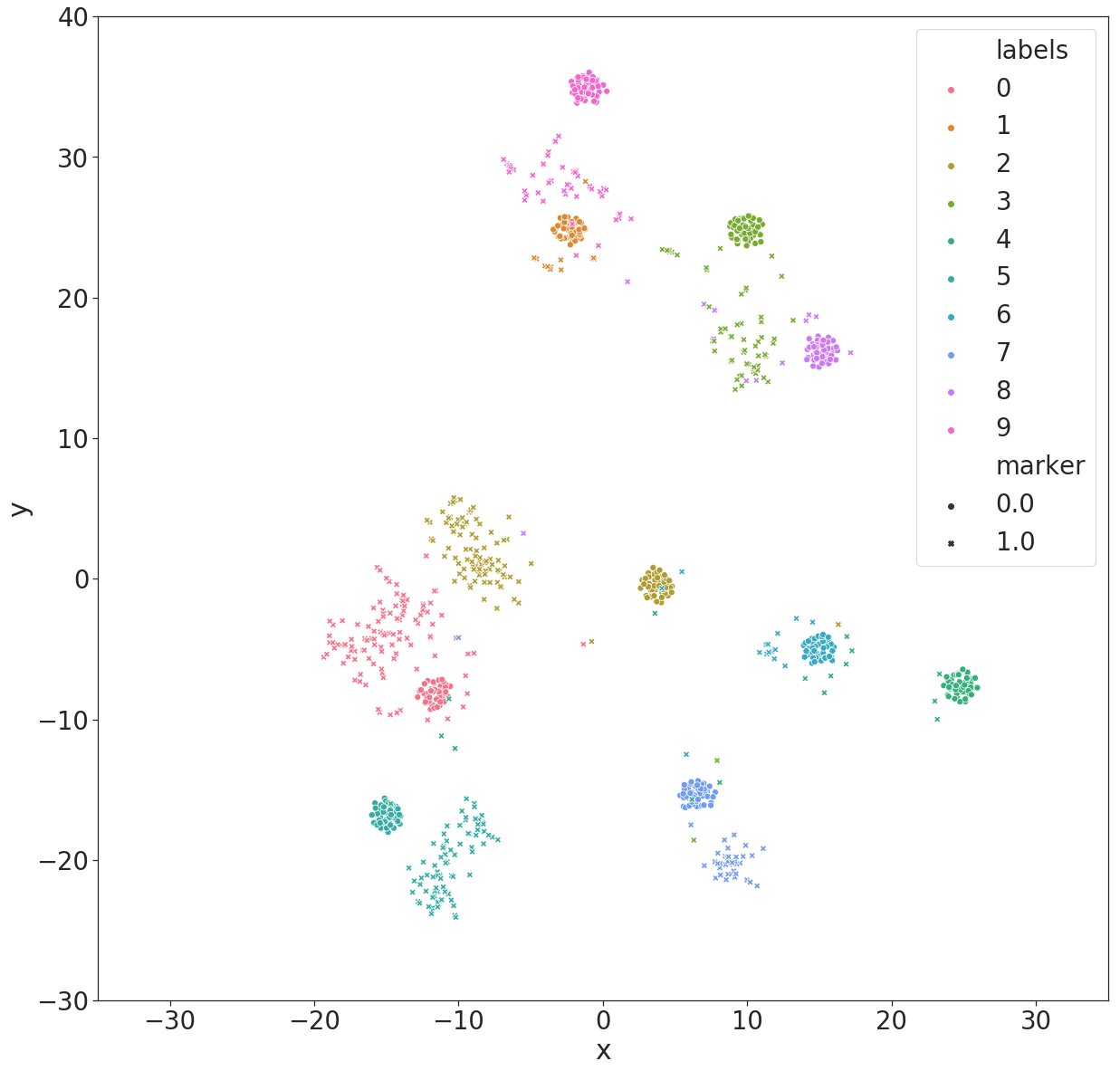} &
    \includegraphics[width=0.32\textwidth,height=0.30\textwidth]{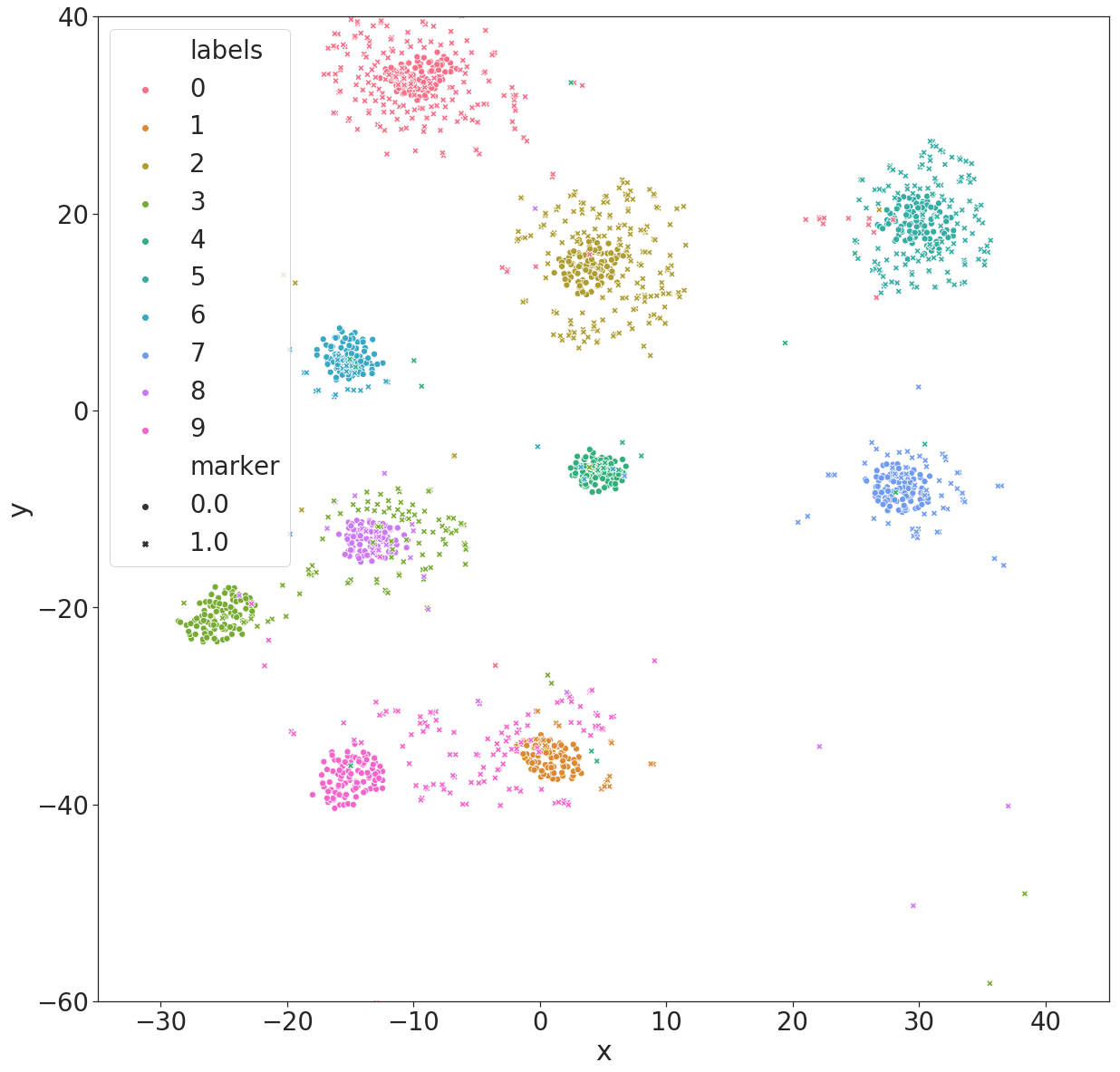} &
    \includegraphics[width=0.32\textwidth,height=0.30\textwidth]{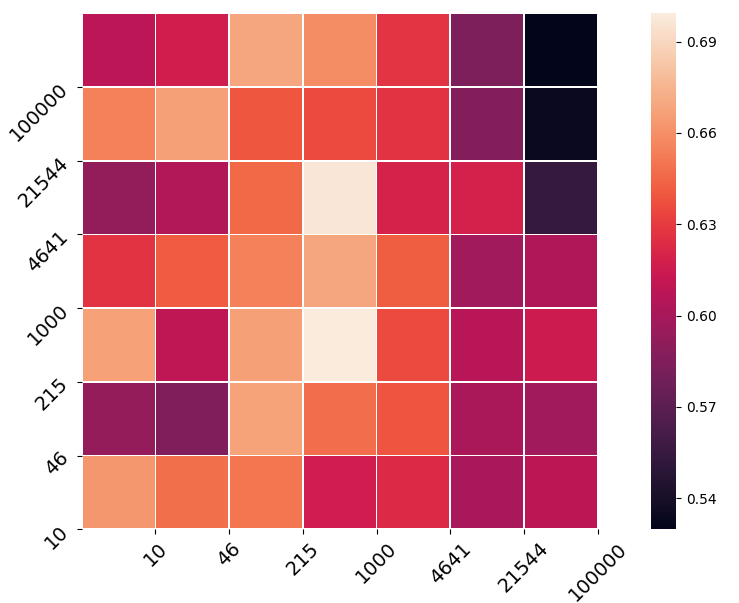}\\
    (a) & (b) &(c)
    \end{tabular}

    \caption{\textbf{(a)} shows the t-SNE plot for the output of the generative model as compared to the test data. Crosses represent the test data while dots represent the generated data. The domain shift is visible in this plot. \textbf{(b)} shows the t-SNE plot after domain shift minimization with our model. The scale for the axes in \textbf{(a)} and \textbf{(b)} is kept constant for comparison. The model can allot the clusters correctly except those where the prediction of pseudo-labels suffered a lot and recovery was difficult. \textbf{(c)} shows the stability of the generative model wrt regularization coefficients on the AWA2 dataset. The x-labels and y-labels are the weight decay in Adam optimizer for learning the NN parameters predicting the Mean and Sigma of the class-conditional distributions respectively. The shaded grid values represent the top-1 accuracy obtained for the given configuration of hyperparameters. Note that even on a logarithmic scale the changes in accuracy are about 1-3\%}
    \label{fig:Clustering}
\end{figure*}
To demonstrate the effectiveness of the proposed approach we performed extensive experimentation on the three standard datasets for ZSL, namely AWA2 \cite{xian2018zero}, CUB-200 \cite{welinder2010caltech} and SUN \cite{xiao2010sun}. In all the experiments, we follow the newly proposed train test split suggested by \cite{xian2018zero}. Since we are using the pre-trained resnet-101 model, therefore, we first sought to make sure that any class that belongs to the test classes is not present in the ImageNet \cite{imagenet2015} training samples. This was already rectified in the split proposed by \cite{xian2018zero} for ZSL. For reference, the network architecture and training procedure is provided in the supplementary material.
\begin{table}[H]
\small
\centering
 \addtolength{\tabcolsep}{1.0pt}
 \begin{tabular}{||c|c|c|c||} 
 \hline
 Dataset & Attribute/Dim & \#Image & Seen/Unseen Class \\ [0.5ex] 
 \hline\hline
AWA2 & A/85 & 37322 & 40/10 \\
CUB & A/1024 & 11788 & 150/50 \\
SUN & A/102 & 14340 & 645/72 \\
 \hline
 \hline
 \end{tabular}
 \caption{Datasets used in our experiments, and their statistics} \label{tab:data}
 \vspace{-2em}
\end{table}
\textbf{Animal with Attribute (AWA2):} The dataset has 50 classes of animals, with 40 classes used for training data and the rest 10 as test data. Each class also has a human annotated 85-dimensional attribute vector associated with it.

\textbf{CUB-200:} This is a fine-grained dataset, containing 200 classes of birds, with 150/50 as the train/test ZSL class split. It has $11788$ data points and 1024-dimensional human-annotated attribute vectors for each class. The attribute vector comprises of 312-dimensional CUB vector appended with word vectors of class names as proposed by \cite{xian2018zero}.

\textbf{SUN Seen Recognition:} There are a total of 14340 images from 717 classes. Hence, every class has nearly 20 samples. Each class is associated with a 102-dimensional human-annotated attribute vector.

\subsection{Zero-Shot Learning (ZSL)}


We report per class accuracy as is the convention in standard ZSL. It is a better metric to report the accuracy of the model as compared to the overall (across classes) accuracy when the classes are unbalanced. We use the newly proposed splits \cite{xian2018zero} for dividing the train and test examples to ensure that the Imagenet classes (used for training ResNet) and the test classes are disjoint. We use the corresponding attribute vector provided against each dataset. Please refer to table-\ref{tab:data} for details on the dataset.

The results of the ZSL setting are shown in the table-\ref{tab:zsl}. As apparent from the table, the proposed approach shows a significant improvement over the previous state-of-art approaches. On the SUN dataset and the AWA2 dataset, we have our top-1 accuracies $63.3\% $ and $70.4\%$ respectively, which are better than its close competitor \cite{verma2017simple}. Also, their top-1 accuracy on the fine-grained CUB dataset is \textbf{significantly} reduced to $49.2\%$,  compared to our model's top-1 accuracy of $70.9\%$. Our model thus performs consistently well and beats other models on all the three benchmark datasets.

Additionally, our approach is more stable to hyperparameter variations as compared to the other competing generative approaches like GFZSL\cite{verma2017simple}. We get only 2-4\% drops in accuracy on a logarithmic scale, unlike GFZSL\cite{verma2017simple} (\textit{figure \ref{fig:Clustering},c}). ADA model has three hyperparameters $\xi,\chi,\beta$ and we chose $\chi=10,\beta =0.5\chi$  as used by the original Cycle GAN \cite{cyclegan}. Random search was used for $\xi=0.0001$.


\begin{table}[h!]
\small
 \centering
 \addtolength{\tabcolsep}{5pt}
 \begin{tabular}{|l| c |c | c | c |} 
 \hline
  & {\textbf{SUN}} & {\textbf{CUB}} &  {\textbf{AWA2}} \\ \hline
 \textbf{Method} &  \textbf{PS} &   \textbf{PS} &  \textbf{PS} \\ 
  \hline
 \textbf{DAP}\cite{DAP} & 39.9  & 40.0    & 46.1 \\
 \textbf{IAP} \cite{IAP}  & 19.4  & 24.0    & 35.9\\
 \textbf{CONSE} \cite{norouzi2013zero} &  38.8 &  34.3   & 44.5 \\
 \textbf{CMT} \cite{cmt}  & 39.9  & 34.6 &  37.9 \\
  \textbf{SSE} \cite{saligram2016learningJoint}  & 51.5  & 43.9 & 61.1\\
  \textbf{LATEM} \cite{latem}  & 55.3  & 49.3    & 55.8\\
 \textbf{DEVISE} \cite{frome2013devise}  & 56.5 & 52.0 & 59.7 \\
 \textbf{SJE} \cite{SJE}  & 53.7  & 53.9 &  61.9 \\
 \textbf{ESZSL} \cite{ESZSL2015} & 54.5  & 53.9   & 58.6 \\
 \textbf{SYNC}\cite{changpinyo2016synthesized}  & 56.3  & 55.6  & 46.6\\
 \textbf{SAE} \cite{SAE2017}  & 40.3  & 33.3  & 54.1 \\
 \textbf{DEM} \cite{dem}  & 61.9  & 51.7  &  67.1 \\

\textbf{GFZSL}\cite{verma2017simple}  & 63.1  & 49.2   & 67.0 \\
 \textbf{CVAE-ZSL}\cite{mishra2017generative}  & 61.7  & 52.1  &  65.8 \\
 \hline
 \textbf{W/O ADA (Ours)} & \textbf{63.3} & \textbf{70.9}  &  \textbf{70.4} \\
 \hline
 \end{tabular}
 \caption{Zero Shot Learning Accuracy on the SUN, CUB, and AWA2 dataset. Here PS is the proposed split recently adopted in the ZSL community after \cite{xian2018zero}. The results reported in the table for the other approaches were taken from the Table 3 of \cite{xian2018zero} }
 \label{tab:zsl} 
 \vspace{-1em}
\end{table}
\begin{table}[h!]
\small
 \centering
 \addtolength{\tabcolsep}{7pt}
 \begin{tabular}{|l| c |c | c | c |} 
 \hline
 \textbf{Method}  & {\textbf{SUN}} & {\textbf{CUB}} &  {\textbf{AWA2}} \\ 
  \hline   
 \textbf{DSRL}\cite{ye2017zero_DOMAIN} & 56.8 & 48.7 & 72.8 \\
 \textbf{ALE} \cite{akata2013label}  & 55.7 & 54.5 & 70.7\\
 \textbf{GFZSL} \cite{verma2017simple} &  64.2 &  50.5   & \textbf{78.6} \\
  \hline
 \textbf{With ADA (Ours)} & \textbf{65.5}  & \textbf{74.2} &  \textbf{78.6} \\
 \hline
 \end{tabular}
 \caption{Transductive Zero-Shot Learning results on the SUN, CUB, and AWA2 dataset. Transductive setting for our model corresponds to ADA. We note that the compared results are reported using the same ResNet101 feature and same train-test split. The results are taken from~\cite{xian2018zero} paper which has evaluated the models with ResNet101 features. }
 \label{tab:adversarialzsl}
 \vspace{-1.5em}
\end{table}

\subsection{Domain Adaption}
In ZSL, since $S\cap U=\phi$, there is a high probability that the seen and 
unseen data do not come from the same underlying domain. This implies that the estimated parameters for the unseen classes, based on the training data of the seen class are likely to deviate from their optimal values. To this end, we propose an Adversarial Domain Adaptation (ADA) method (refer section \ref{sec:ADA}) to explicitly handle the domain shift problem. 

In Table-\ref{tab:adversarialzsl}  we show the results of the proposed ADA method and compare against the previous transductive setting approaches. The result of ALE \cite{akata2013label} and GFZSL \cite{verma2017simple} are taken from the Figure 8 of \cite{xian2018zero}. Here we observe that using the domain adaption method boosts the generative model's performance. In the case of the AWA2 dataset without domain adaption, the top-1 accuracy was 70.4\% while with the domain adaption it rises to 78.6\%. A similar pattern is observed for the CUB (3.3\% improvement) and SUN dataset also. The domain shift in SUN dataset is ameliorated by the presence of a large number of training and testing classes and hence we see a smaller increment after ADA. 

Moreover, our ADA method can minimize the domain shift (apparent in the Figure \ref{fig:Clustering} (a),(b)) in accordance with the clusters allotted by the base ZSL model. We can see that the model associates wrong clusters for only two classes owing to the low prediction accuracy of the base ZSL model for these classes which, itself is due to a strong overlap in test clusters of these classes. Thus, a reduction in label corruption will further improve the domain matching.




\subsection{Ablation Study}

In this section, we compare variants of our proposed approach through an ablation study to empirically analyze the benefits of each component. In particular, we check whether enforcing cyclic consistency leads to better performance than the vanilla adversarial loss, whether incorporating deep classifiers in architecture leads to improved performance, and whether adversarial domain adaptation is required for domain shift minimization for training the deep classifiers. 
\subsubsection{Experimental Setup}
We have kept the pre-trained base ZSL model the same for consistent ablation results. The different variants of our model for the ablation study are described below:

\begin{itemize}
    \item \textbf{Std DA}: To test the relative importance of adversarial domain adaptation and hence domain shift minimization, we trained a deep classifier (with the same architecture as other variants) on the labeled samples synthesized from our generative model (base ZSL model) and the unlabelled test data with its pseudo labels.
    \item \textbf{Vanilla ADA}: This domain adaptation model comprises of a single generator and discriminator augmented with a classifier, where the generator maps the source domain to the target domain. For effective comparison, we used the same architecture of generators, discriminators and classifiers for ablation and experimental evaluations.  
    \item \textbf{CycleGAN w/o}: In this variant, we removed the classifiers $C^T$ and $C^S$ associated with our proposed ADA model. Hence, the adversarial architecture is similar to CycleGAN which comprises of two generators and associated two discriminators.
    \item \textbf{Ours}: This is our proposed ADA model, defined in section \ref{sec:ADA}

\end{itemize}

We employ two different techniques for predicting the class labels. For the above-defined variants which have a trainable classifier in them like \textit{vanilla ADA} and \textit{Ours}, we report the class averaged top-1 accuracy of the predictions from the classifier attached to the discriminator (referred as \textbf{M1} in \textit{table \ref{tab:ablationzsl}}). For the approach \textit{Ours}, classifier $C^T$ ( mapping the source domain to the target domain ) is used for class label predictions.

We also report the 1 nearest neighbor classification accuracies using the Gaussian distance between the class conditionals mapped to the target domain by the generator and the test data feature (referred to as \textbf{M2} in Table \ref{tab:ablationzsl}). This method predicts the most probable class via the mixture of class conditionals, in a similar way like the base ZSL model in the inductive setting. To generate the mapped cluster prototypes for each class conditional, we sample the data points from its class conditional distribution, transform them into another target domain (test domain) via the generator of ADA and then extract the required statistics (mean of the cluster for our case) from the new distribution. Like ADA experimental setup, the learned covariance matrix is not changed after domain adaptation. 

For the variant without a trainable classifier, \textit{CycleGAN w/o}, we only use the later method (method M2)  for evaluating the accuracies. Also, note that due to the absence of any adversarial generator in the variant \textit{Std DA}, the \textbf{M2} accuracy is computed in the exact same way as in our base inductive ZSL model.
\begin{table}[h!]
\scriptsize
 \centering
 \addtolength{\tabcolsep}{0.2pt}
 \begin{tabular}{|l| c | c | c | c | c | c |} 
 \hline
   & \multicolumn{2}{c|}{\textbf{SUN}} & \multicolumn{2}{c|}{\textbf{CUB}} &  \multicolumn{2}{c|}{\textbf{AWA2}} \\
  \hline
  \textbf{Variant} & \textbf{M1}& \textbf{M2}& \textbf{M1} & \textbf{M2} & \textbf{M1} & \textbf{M2} \\
  \hline
 \textbf{Std DA}  & 64.8  & NA & 72.2 & NA & 71.3 & NA\\
 \hline
 \textbf{Vanilla ADA}  & 64.9  & 47.1 & 71.5 & 57.8 & 77.3 & 56.1\\
 \hline
 \textbf{CycleGAN w/o}  & NA  &  \textbf{57.2}  & NA & \textbf{68.4} & NA & \textbf{75.8}\\
  \hline
 \textbf{Ours} & \textbf{65.5}  & {55.8} & \textbf{74.2}  & 67.5 & \textbf{78.6} & 74.9\\
 \hline
 \end{tabular}
 \caption{Ablation study on ZSL with splits proposed in \cite{xian2018zero}}
 \label{tab:ablationzsl}
 \vspace{-3.5em}
\end{table}

\subsubsection{Analysis}
When we compare the performance of \textit{Std DA} with the base inductive ZSL model (results in Table \ref{tab:zsl}), we only see a marginal performance increase. During the experiments, the classifiers initially came close to our benchmark model (\textit{Ours}), but soon converged to a sub-optimum where they mimicked the accuracy of pseudo-labels provided by the base ZSL model. Owing to the domain shift, the classifier was not able to transfer the supervision from generated samples to the test data. This supports the claim that adversarial networks reduce domain mismatch, precluding the classifiers from converging at pseudo-labels. 

The addition of trainable classifiers with ADA gave a heavy accuracy boost. This is mostly because of the higher expressivity and generalizability of such neural net classifiers as compared to nearest neighbor based classifiers. This is empirically suggested by diminished performance of about 3-10\% on various datasets in \textit{CycleGan w/o} wrt \textbf{M1} accuracy of \textit{Ours}. The addition of classification loss term does reduce the linear separability (reduction in \textbf{M2} score of \textit{Ours} vs \textit{CycleGAN w/o}) but the performance gain from classifiers overshadows this degradation.

Cyclic consistency further restrains the output space of the generator which drastically improves the linear separability of the generated data points (\textbf{M2} score of \textit{Ours}). This causes the proposed model to perform better than standard adversarial architecture using a similar classifier. This is apparent when we compare \textbf{M2} accuracies of \textit{vanilla ADA} with \textit{ours}. Even though the \textbf{M1} accuracies of these two models differ by about 1-2\%, the drop in \textbf{M2} accuracies are severe. Since nearest neighbor models rely on linear separability they suffer with as large as 10-30\% drop.This is also apparent in \textit{figure \ref{fig:Clustering}} t-SNE plots.

We can safely conclude, adding adversarial domain adaptation to the generative ZSL framework allows us to leverage the expressivity of neural net classifiers to classify novel classes while being trained only using the labels from seen classes. The adversarial adaptation minimizes the domain shift which is a crucial requirement for classifiers to transfer knowledge from the synthesized data and hence helps to train incisive classifiers that do not face the hubness issue, unlike distance-based nearest neighbor classifiers.

\section{Conclusion}
In this paper, we address the issue of domain shift between the distributions of the seen and unseen classes in zero-shot learning. We adopt an end-to-end approach for generative modeling that captures non-linear dynamics better as compared to previous state-of-the-art approaches. The proposed approach first learns the class conditional distributions for both the seen and unseen classes by leveraging the data from only the seen classes. Following this, we explicitly minimize the domain shift between the estimated unseen class distributions and the true unseen class distributions by using a cyclic consistency based adversarial scheme. We show through detailed experimentation, that our proposed generative model, although much simpler than GAN/VAE based frameworks, outperform existing models in the ZSL setting. Also, we show that our scheme of minimizing domain shift significantly improves performance, as compared to the transductive setting methods adopted by previous approaches. The generative framework can in principle assume any form, some of the popular ones being GAN and VAE based models. However, they lack explainability and they require further sampling to extract statistics like class variance. A larger intra-class variance would be an outcome of larger variations in the visual appearances of class attributes and hence the samples would be harder to classify together. An interesting future direction can be to use these statistics to model selective attention mechanisms or training with hard negative mining.

\textbf{Acknowledgements:} VKV acknowledges support from Visvesvaraya PhD Fellowship and PR acknowledges support from Visvesvaraya Young Faculty Fellowship.

{\small
\bibliographystyle{ieee}
\bibliography{wacv_zsl}

\begin{thebibliography}{10}\itemsep=-1pt

\bibitem{akata2013label}
Z.~Akata, F.~Perronnin, Z.~Harchaoui, and C.~Schmid.
\newblock Label-embedding for attribute-based classification.
\newblock In {\em CVPR}, pages 819--826, 2013.

\bibitem{SJE}
Z.~Akata, S.~Reed, D.~Walter, H.~Lee, and B.~Schiele.
\newblock Evaluation of output embeddings for fine-grained image
  classification.
\newblock In {\em Proceedings of the IEEE Conference on CVPR}, pages
  2927--2936, 2015.

\bibitem{arjovsky2017wasserstein}
M.~Arjovsky, S.~Chintala, and L.~Bottou.
\newblock Wasserstein gan.
\newblock {\em arXiv preprint arXiv:1701.07875}, 2017.

\bibitem{recgan}
H.~Bharadhwaj, H.~Park, and B.~Y. Lim.
\newblock Recgan: recurrent generative adversarial networks for recommendation
  systems.
\newblock In {\em Proceedings of the 12th ACM Conference on Recommender
  Systems}, pages 372--376. ACM, 2018.

\bibitem{icra}
H.~Bharadhwaj, Z.~Wang, Y.~Bengio, and L.~Paull.
\newblock A data-efficient framework for training and sim-to-real transfer of
  navigation policies.
\newblock {\em arXiv preprint arXiv:1810.04871}, 2018.

\bibitem{BucherZSL}
M.~Bucher, S.~Herbin, and F.~Jurie.
\newblock Generating visual representations for zero-shot classification.
\newblock In {\em ICCV Workshops}, Oct 2017.

\bibitem{changpinyo2016synthesized}
S.~Changpinyo, W.-L. Chao, B.~Gong, and F.~Sha.
\newblock Synthesized classifiers for zero-shot learning.
\newblock In {\em Proceedings of the IEEE Conference on Computer Vision and
  Pattern Recognition}, pages 5327--5336, 2016.

\bibitem{Chen_2018_CVPR}
L.~Chen, H.~Zhang, J.~Xiao, W.~Liu, and S.-F. Chang.
\newblock Zero-shot visual recognition using semantics-preserving adversarial
  embedding networks.
\newblock In {\em The IEEE Conference on Computer Vision and Pattern
  Recognition (CVPR)}, June 2018.

\bibitem{frome2013devise}
A.~Frome, G.~S. Corrado, J.~Shlens, S.~Bengio, J.~Dean, T.~Mikolov, et~al.
\newblock Devise: A deep visual-semantic embedding model.
\newblock In {\em NIPS}, pages 2121--2129, 2013.

\bibitem{GAN}
I.~Goodfellow, J.~Pouget-Abadie, M.~Mirza, B.~Xu, D.~Warde-Farley, S.~Ozair,
  A.~Courville, and Y.~Bengio.
\newblock Generative adversarial nets.
\newblock In {\em Advances in neural information processing systems}, pages
  2672--2680, 2014.

\bibitem{guo2017synthesizing}
Y.~Guo, G.~Ding, J.~Han, and Y.~Gao.
\newblock Synthesizing samples for zero-shot learning.
\newblock In {\em IJCAI}, 2017.

\bibitem{resnet}
K.~He, X.~Zhang, S.~Ren, and J.~Sun.
\newblock Deep residual learning for image recognition.
\newblock In {\em CVPR}, pages 770--778, 2016.

\bibitem{hinton2012neural}
G.~Hinton, N.~Srivastava, and K.~Swersky.
\newblock Neural networks for machine learning lecture 6a overview of
  mini-batch gradient descent.
\newblock {\em Cited on}, 14:8, 2012.

\bibitem{cycada}
J.~Hoffman, E.~Tzeng, T.~Park, J.-Y. Zhu, P.~Isola, K.~Saenko, A.~A. Efros, and
  T.~Darrell.
\newblock Cycada: Cycle-consistent adversarial domain adaptation.
\newblock {\em arXiv preprint arXiv:1711.03213}, 2017.

\bibitem{kingma2014adam}
D.~Kingma and J.~Ba.
\newblock Adam: A method for stochastic optimization.
\newblock {\em arXiv preprint arXiv:1412.6980}, 2014.

\bibitem{VAE}
D.~P. Kingma and M.~Welling.
\newblock Auto-encoding variational bayes.
\newblock {\em ICLR}, 2014.

\bibitem{UDA}
E.~Kodirov, T.~Xiang, Z.~Fu, and S.~Gong.
\newblock Unsupervised domain adaptation for zero-shot learning.
\newblock In {\em Proceedings of the IEEE International Conference on Computer
  Vision}, pages 2452--2460, 2015.

\bibitem{kodirov2015unsupervisedDA}
E.~Kodirov, T.~Xiang, Z.~Fu, and S.~Gong.
\newblock Unsupervised domain adaptation for zero-shot learning.
\newblock In {\em ICCV}, pages 2452--2460, 2015.

\bibitem{SAE2017}
E.~Kodirov, T.~Xiang, and S.~Gong.
\newblock Semantic autoencoder for zero-shot learning.
\newblock {\em arXiv preprint arXiv:1704.08345}, 2017.

\bibitem{DAP}
C.~H. Lampert, H.~Nickisch, and S.~Harmeling.
\newblock Attribute-based classification for zero-shot visual object
  categorization.
\newblock {\em IEEE Transactions on PAMI}, 36(3):453--465, 2014.

\bibitem{IAP}
C.~H. Lampert, H.~Nickisch, and S.~Harmeling.
\newblock Attribute-based classification for zero-shot visual object
  categorization.
\newblock {\em IEEE Transactions on PAMI}, 36(3):453--465, 2014.

\bibitem{lu2017zero}
J.~Lu, J.~Li, Z.~Yan, and C.~Zhang.
\newblock Zero-shot learning by generating pseudo feature representations.
\newblock {\em arXiv preprint arXiv:1703.06389}, 2017.

\bibitem{mishra2017generative}
A.~Mishra, M.~Reddy, A.~Mittal, and H.~A. Murthy.
\newblock A generative model for zero shot learning using conditional
  variational autoencoders.
\newblock {\em arXiv preprint arXiv:1709.00663}, 2017.

\bibitem{norouzi2013zero}
M.~Norouzi, T.~Mikolov, S.~Bengio, Y.~Singer, J.~Shlens, A.~Frome, G.~S.
  Corrado, and J.~Dean.
\newblock Zero-shot learning by convex combination of semantic embeddings.
\newblock {\em arXiv preprint arXiv:1312.5650}, 2013.

\bibitem{conGAN}
S.~E. Reed, Z.~Akata, X.~Yan, L.~Logeswaran, B.~Schiele, and H.~Lee.
\newblock Generative adversarial text to image synthesis.
\newblock {\em CoRR}, abs/1605.05396, 2016.

\bibitem{ESZSL2015}
B.~Romera, Paredes and P.~Torr.
\newblock An embarrassingly simple approach to zero-shot learning.
\newblock In {\em International Conference on Machine Learning}, pages
  2152--2161, 2015.

\bibitem{imagenet2015}
O.~Russakovsky, J.~Deng, H.~Su, J.~Krause, S.~Satheesh, S.~Ma, Z.~Huang,
  A.~Karpathy, A.~Khosla, M.~Bernstein, et~al.
\newblock Imagenet large scale visual recognition challenge.
\newblock {\em IJCV}, pages 211--252, 2015.

\bibitem{cmt}
R.~Socher, M.~Ganjoo, C.~D. Manning, and A.~Ng.
\newblock Zero-shot learning through cross-modal transfer.
\newblock In {\em NIPS}, pages 935--943, 2013.

\bibitem{song2018transductive}
J.~Song, C.~Shen, Y.~Yang, Y.~Liu, and M.~Song.
\newblock Transductive unbiased embedding for zero-shot learning.
\newblock In {\em Proceedings of the IEEE Conference on Computer Vision and
  Pattern Recognition}, pages 1024--1033, 2018.

\bibitem{robustGAN}
K.~K. Thekumparampil, A.~Khetan, Z.~Lin, and S.~Oh.
\newblock Robustness of conditional gans to noisy labels.
\newblock In S.~Bengio, H.~Wallach, H.~Larochelle, K.~Grauman, N.~Cesa-Bianchi,
  and R.~Garnett, editors, {\em Advances in Neural Information Processing
  Systems 31}, pages 10271--10282. Curran Associates, Inc., 2018.

\bibitem{adda}
E.~Tzeng, J.~Hoffman, K.~Saenko, and T.~Darrell.
\newblock Adversarial discriminative domain adaptation.
\newblock In {\em CVPR}, 2017.

\bibitem{vermageneralized}
V.~K. Verma, G.~Arora, A.~Mishra, and P.~Rai.
\newblock Generalized zero-shot learning via synthesized examples.
\newblock {\em CVPR}, 2018.

\bibitem{verma2017simple}
V.~K. Verma and P.~Rai.
\newblock A simple exponential family framework for zero-shot learning.
\newblock In {\em ECML-PKDD}, pages 792--808. Springer, 2017.

\bibitem{irgan}
J.~Wang, L.~Yu, W.~Zhang, Y.~Gong, Y.~Xu, B.~Wang, P.~Zhang, and D.~Zhang.
\newblock Irgan: A minimax game for unifying generative and discriminative
  information retrieval models.
\newblock In {\em Proceedings of the 40th International ACM SIGIR conference on
  Research and Development in Information Retrieval}, pages 515--524. ACM,
  2017.

\bibitem{wang2017zero}
W.~Wang, Y.~Pu, V.~K. Verma, K.~Fan, Y.~Zhang, C.~Chen, P.~Rai, and L.~Carin.
\newblock Zero-shot learning via class-conditioned deep generative models.
\newblock {\em arXiv preprint arXiv:1711.05820}, 2017.

\bibitem{welinder2010caltech}
P.~Welinder, S.~Branson, T.~Mita, C.~Wah, F.~Schroff, S.~Belongie, and
  P.~Perona.
\newblock Caltech-ucsd birds 200.
\newblock 2010.

\bibitem{latem}
Y.~Xian, Z.~Akata, G.~Sharma, Q.~Nguyen, M.~Hein, and B.~Schiele.
\newblock Latent embeddings for zero-shot classification.
\newblock In {\em Proceedings of the IEEE Conference on Computer Vision and
  Pattern Recognition}, pages 69--77, 2016.

\bibitem{xian2018zero}
Y.~Xian, C.~H. Lampert, B.~Schiele, and Z.~Akata.
\newblock Zero-shot learning-a comprehensive evaluation of the good, the bad
  and the ugly.
\newblock {\em IEEE transactions on pattern analysis and machine intelligence},
  2018.

\bibitem{xian2018feature}
Y.~Xian, T.~Lorenz, B.~Schiele, and Z.~Akata.
\newblock Feature generating networks for zero-shot learning.
\newblock In {\em Proceedings of the IEEE Conference on Computer Vision and
  Pattern Recognition}, 2015.

\bibitem{xiao2010sun}
J.~Xiao, J.~Hays, K.~A. Ehinger, A.~Oliva, and A.~Torralba.
\newblock Sun database: Large-scale scene recognition from abbey to zoo.
\newblock In {\em CVPR, 2010}, pages 3485--3492. IEEE, 2010.

\bibitem{ye2017zero_DOMAIN}
M.~Ye and Y.~Guo.
\newblock Zero-shot classification with discriminative semantic representation
  learning.
\newblock In {\em CVPR}, 2017.

\bibitem{dem}
L.~Zhang, T.~Xiang, S.~Gong, et~al.
\newblock Learning a deep embedding model for zero-shot learning.
\newblock 2017.

\bibitem{saligram2016learningJoint}
Z.~Zhang and V.~Saligrama.
\newblock Learning joint feature adaptation for zero-shot recognition.
\newblock {\em arXiv preprint arXiv:1611.07593}, 2016.

\bibitem{cyclegan}
J.-Y. Zhu, T.~Park, P.~Isola, and A.~A. Efros.
\newblock Unpaired image-to-image translation using cycle-consistent
  adversarial networks.

\end{thebibliography}
}

\clearpage
\newpage
\appendix

\section{Supplementary: Implementation Details}
\subsection{Generative Framework}
In this section, we describe the architecture that yields our reported results in the ZSL setting wherein there are no images from seen classes in test samples. For the SUN dataset, both the networks used for modeling mean and co-variance have linear (1800 and 2048 nodes), batch normalization and Relu layers. Additionally, the co-variance is restricted to be in the range of 0.5 to 1.5 for numerical stability via sigmoid activation. Both the networks are trained with ADAM optimizer \cite{kingma2014adam} using 0.001 and 0.1 as regularizer coefficients for means and covariance respectively.

For the AWA dataset, the generator networks have an architecture similar to SUN but consist of an additional dropout layer with probability 0.1. The parameters of the means are regularized with the coefficient $10^3$ while the parameters of covariance are regularized with the coefficient $10^4$.

For the CUB dataset, the generator networks have three linear layers (1200, 1800, 2048 nodes), 2 Relu, 2 batch normalization and 2 dropout layers. Their regularization coefficients are 0.01 and 0.1 for mean and covariance respectively. All the above networks are trained with a learning rate of $0.00001$. The training of these networks was additionally regularized via early stopping
\subsection{Adversarial Domain Adaption} As described above, ADA model comprises two discriminators ($D^{S,T}$), two classifiers ($C^{S,T}$) and two generators ($G^{S,T}$). 

The discriminators $D^{S,T}$ are 5 layered neural networks comprising of 2 Linear layers of 1600 nodes and 1 node respectively, a single leaky Relu layer with a negative slope of 0.2 and batch normalization. The classifiers are single-layered networks with the number of nodes equal to the number of classes. $log(softmax(\mathbf{x}))$ is used as activation function for the classifiers $C^{S,T}$. The generators $G^{S,T}$ consist of three linear layers (1200,1200 and 2048 nodes), dropout layers, batch normalization and leaky Relu.

The overall objective is minimized using RMSprop (\cite{hinton2012neural}) optimizer with a learning rate of 0.00001. A manual seed of 100 has been used for all the ADA experiments.

\section{Supplementary: Training Procedures}
For brevity, we provide the training algorithms for our base ZSL model and the ADA ZSL model. We use the following loss definition for adversarial training as described in the paper
\begin{equation}
  \mathcal{L}= \mathcal{L}_{adv}^T+\mathcal{L}_{adv}^S+\chi \mathcal{L}_{cyc}+\xi\mathcal{L}_{clf}^T + \xi\mathcal{L}_{clf}^S  
\end{equation}
where $\mathcal{L}_{adv}^{\{T,S\}} = \{L_G +L_D\}^{\{T,S\}}$ with
\begin{equation}
    L_G^{T} = \underset{c \sim {p_{c}}}{\mathbb{E}}[\beta \norm{G^T(x_{nc})-x_{nc}}_p - D^T_w \circ G^{T}(y_{nc})]
\end{equation}
\begin{equation}
    L_G^{S} = \underset{c \sim {p_{c}}}{\mathbb{E}}[\beta \norm{G^S(y_{nc})-y_{nc}}_p - D^S_w \circ G^{S}(x_{nc})]
\end{equation}
\begin{equation}
L^{T}_D =  \mathbb{E}_{c \sim {p_{c}}}[D^{T}_w \circ G^{T}(y_{nc})] - \mathbb{E}_{c \sim {p_{c}}}[D^T_w({x_{nc}})] 
\end{equation}
\begin{equation}
L^{S}_D =  \mathbb{E}_{c \sim {p_{c}}}[D^{S}_w \circ G^{S}(x_{nc})] - \mathbb{E}_{c \sim {p_{c}}}[D^S_w(y_{nc})] 
\end{equation}
Here, $D_w$ is the Wasserstein loss~\cite{arjovsky2017wasserstein} and $c$ denotes the class label.
\begin{align}
& \mathcal{L}_{\text{cyc}}(G^T, G^S)  =  \mathbb{E}_{c\sim p_{\text{c}}}[\norm{G^{S}\circ G^T(y_{nc})- x_{nc}}_p] \nonumber \\  &
+  \mathbb{E}_{c\sim p_{\text{c}}}[\norm{G^{T}\circ G^S(x_{nc})- y_{nc}}_p].\lbleq{cycle}
\end{align}
Here, $||\cdot||_p$ denotes the $L_p$ norm.
\begin{align}
L^{T}_{clf} = \mathbb{E}_{c \sim {p_{c}}}[L(C^{T}_{clf}\circ G^{T}(y_{nc}),Y^{T})] \nonumber \\  
+ \mathbb{E}_{c \sim {p_{c}}}[L(C^{T}_{clf}(x_{nc}),\bar{Y}^U)]
\end{align}
\begin{align}
L^{S}_{clf} = \mathbb{E}_{{c} \sim {p_{c}}}[L(C^{S}_{clf}\circ G^{S}(x_{nc}),Y^{S})] \nonumber \\  
+ \mathbb{E}_{c \sim {p_{c}}}[L(C^{S}_{clf}(y_{n c}),\bar{Y}^U)]
\end{align}

\begin{algorithm}[H]
	\caption{\textsc{Superficial Training Scheme} \label{alg:system}}
	\begin{algorithmic}[1]
		\State Train ResNet101 on Imagenet
		\State Randomly initialize model parameters
		\State Initialize dataset $\mathcal{D}_{ada}\gets \emptyset $
			\State Train the generative model to describe data from class $c$ by $p(\mathbf{x}|c,\Theta)$ $\forall c\in[1,...C]$
			\State $Source \longleftarrow$ data sampled from generative model
			\State $Target \longleftarrow$ unlabelled test data
			\State Initialize weights of $G^T:S\longrightarrow T$,$G^S:T\longrightarrow S$, $D^S$, $D^T$ 
			\State Augment class labels to $G^T$ and $G^S$
		\For{epoch $i=1:K$}
		\State Randomly sample $o_i^{s}\sim Source$, $o_i^{t}\sim {Target}$ 
		\State Perform ADA
		\EndFor
	
	\end{algorithmic}
\end{algorithm}
\begin{algorithm}[H]
	\begin{algorithmic}[1]
		\State {\bfseries Input:}Maximum iterations $N_{step}$, batch size $n$, the iteration number of discriminator in a loop $n_{d}$, RMSprop hyperparameters $\alpha$, class attributes $\{a_c\}_{c=1}^{S+U}$, ADAM hyperparameters $\alpha_2,\beta_1,\beta_2$
		
		\State // \textit{Pre-training}
		\For{$c = 1...S+U$}
		\State $\mathbf{\zeta_c} = f_{\mathbf{\Theta}}(\mathbf{a_c})$
         \EndFor
         \For{iter $= 1,..., N_{step}$}
            \State Sample minibatch of examples $x_1,x_2,...x_n$

         \State
         $L =\smashoperator[r]{\sum\limits_{c=1}^S}\smashoperator[r]{\sum\limits_{n:y_n=c}}\left(x_n - f_{\mathbf{\mu}}(a_c)\right)^T\left[f_{\Sigma}(a_c) \right] \left(x_n - f_{\mathbf{\mu}}(a_c)\right)$
          
          \State  $\mathbf{\Theta} \gets \text{Adam}(\bigtriangledown_{\mathbf{\Theta}} L,\mathbf{\Theta}, \alpha_2, \beta_1, \beta_2)$
         \EndFor
    \State // \textit{Adversarial Domain Adaptation}     
	\State Initialize $\{G,D,C\}^{T,S}$
	\State $\lambda_d = Params(D^T,D^S,C^T,C^S)$
	\State $\lambda_g = Params(G^T,G^S)$
        \For{iter $= 1,..., N_{step}$}
            \State Sample minibatch $x_1,x_2,...x_n$ from test data for training $G^T$
            \State Sample minibatch $x'_1,x'_2,...x'_n$ from class conditional distributions for training $G^S$
        	\State Compute the overall loss $\mathcal{L}$ using Eq.1
        \State $\lambda_g \gets \text{RMSprop}(\bigtriangledown_{\lambda_g} \mathcal{L},\lambda_g, \alpha)$
		\For{$t = 1$, ..., $n_{d}$}
		 \State Sample minibatch of examples $x$
        \State Compute the overall loss $\mathcal{L}$ using Eq.1
        \State $\zeta_d \gets \text{RMSprop}(\bigtriangledown_{\lambda_d} \mathcal{L},\lambda_d, \alpha)$
        \EndFor

		\EndFor
	\end{algorithmic}
	\caption{All the notations have same meaning as that in the running text. \textit{Params}() return the parameters of the model in (.)}
	\label{alg_label}
	
\end{algorithm}

\end{document}